# A Classification of Adjectives for Polarity Lexicons Enhancement

Silvia Vázquez, Núria Bel

Universitat Pompeu Fabra
Roc Boronat 138
08018, Barcelona
E-mail: silvia.vazquez@upf.edu, nuria.bel@upf.edu

**Abstract**

Subjective language detection is one of the most important challenges in Sentiment Analysis. Because of the weight and frequency in opinionated texts, adjectives are considered a key piece in the opinion extraction process. These subjective units are more and more frequently collected in polarity lexicons in which they appear annotated with their prior polarity. However, at the moment, any polarity lexicon takes into account prior polarity variations across domains. This paper proves that a majority of adjectives change their prior polarity value depending on the domain. We propose a distinction between domain dependent and domain independent adjectives. Moreover, our analysis led us to propose a further classification related to subjectivity degree: constant, mixed and highly subjective adjectives. Following this classification, polarity values will be a better support for Sentiment Analysis.

## 1. Introduction

The amount of available information on the Internet is continuously growing. Now, surfing the web to find not only news (objective data) but also opinions (subjective data) has became a daily practice for all of us. People search from opinions about the latest political campaign to product reviews before going shopping. Therefore, the markets have realized that, being able to automatically extract the opinions conveyed in these texts, they can better understand the needs of their customers and they can set up more real and competitive commercial strategies.

Therefore, subjective language detection is one of the most important challenges for Sentiment Analysis. To achieve that objective, developing tools that allow the extraction of the appropriate information from these opinionated texts is necessary. One of the tasks in which tool developers spend more time and efforts is the production of language resources used in the first steps of the opinion mining analysis: polarity lexicons. These are compilations of large amounts of words that tend to appear in opinionated texts. Moreover, all of these words are annotated with their prior polarity, that is, with their semantic orientation (i.e. positive or negative). However, opinion words collected in these polarity lexicons usually overlook prior polarity variations across domains. One example of this variation can be seen in the adjective "small". It will be considered by a majority of people as a good characteristic for mobile phones, however if someone ask if having a "small car" is good, some people will answer yes and others will answer no. Moreover, if we also ask about a "small film" probably people won't know how to answer to our question.

In this work we demonstrate the high impact of polarity variation depending on the domain, specifically for adjectives, after the comparison of prior polarity values of 514 adjectives across three different domains. For this purpose, we compile three domain corpora (cars, mobile phones and films) and extracted common adjectives to compare their polarities. All of these units were manually annotated by 5 human annotators with their prior polarity.

As a result, we proved that more than half of these units (67.32%) have a polarity that is domain dependent. Moreover, the annotation task revealed that, for some adjectives, human annotators showed very little agreement in what seems to be a phenomenon related to subjectivity. By measuring this new source of variation with standard deviation metrics, we could classify polarity adjectives in three different categories depending on the prior polarity perceived by the annotators: constant, mixed and highly subjective.

The proof of the high impact of domain polarity variation of particular adjectives is meant to improve the future accuracy of polarity lexicons by turning them into much more reliable resources. Tools could take into account polarity differences between domains but also the existence of several elements that could be considered very unreliable, since they do not have a fixed polarity, or in contrast, reliable constant units, since they always have the same prior polarity in all domains. Thus, polarity lexicons would be annotated in a more accurate way and would therefore improve the efficiency of opinion mining tools.

The paper is organized as follows. In section 2, we present related work in this field. In section 3, we describe the experiments that we carried out. In section 4, we present the analysis of results. Finally, in section 5 we report on the conclusions and future work.

## 2. Related work

In the literature we can find several attempts to produce polarity lexicons in the last years. One of the resources most well-known is SentiWordNet[1] (Esuli and Sebastiani, 2006; Bacianella et al., 2010) in which each WordNet (Miller, 1995) synset has one of three possible values: Objective, Positive or Negative. The basic idea is that different senses of the same term can have different properties related to opinion.

Valitutti (2004) also tried to improve WordNet's potential by adapting it to polarity issues and identifying all of the lexical elements that have a high load of emotional or affective content against the units that do not express emotion or affect.

Another different method also based on WordNet exploration was done in Hu and Liu (2004). They used a very little set of seed adjectives annotated with their prior polarity. With this information, they found the synset of each seed adjective and also its antonyms. The polarity of an adjective that is in the synset of a seed adjective would be the same as the seed adjective. On the other hand, the polarity of an adjective that is in its antonym set would have the opposite.

Vegnaduzzo (2004) presented a method to automatically build dictionaries using only a POS tagged corpus and a set of seed subjective adjectives. He had hypothesized that adjectives more likely to be subjective are those that are more similar in terms of their context to the adjectives in the seed set.

Some of the latest work has been carried out by Rao and Ravichandran (2009) and Velikovich et al. (2010). Rao and Ravichandran used lexical resources such as WordNet and OpenOffice Thesaurus, while Velikovich et al. presented web-based approaches. Both showed how to detect polarity of the subjective words used in opinionated texts through lexical graphs where the polarity of some known nodes is propagated to other unknown nodes through the edges.

However, none of these works takes into account prior polarity variation across domains. Our work proves the impact of this variation and classifies adjectives based upon it.

On the other hand, much of the research on opinion adjectives, in particular, was performed by Wiebe (Hatzivassiloglou and Wiebe, 2000; Wiebe and Wilson, 2002; Riloff and Wiebe, 2003). Here it was shown that the most evident cues of subjectivity are adjectives, more than verbs or n-grams. These units were referred as "strongly subjective clues".

The study of adjectives' behavior, their frequency as well as their weight in opinion texts, is crucial to improve polarity lexicons and the accuracy of analysis tools. Our goal is to prove the high presence of domain dependent elements to demonstrate why polarity lexicons should be adapted to this variability.

## 3. Experiments

To assess the impact of domain variation, we compared adjectives that appeared in different domains. For this purpose, we built three opinion text corpora of different domains and all of the adjectives that appeared in the three domains were annotated by five human annotators. The comparison of the values received (see 3.2) showed that the number of domain dependent adjectives was more than the double of independent ones. Moreover, the comparison of the annotations made by different annotators showed a high variation not only across domains but also in the same domain. Further analysis confirmed a new classification: constant, mixed and highly subjective adjectives.

### 3.1 Corpora

We have built three corpora (300K words each) from three different domains: cars, mobile phones and films. All of the texts were extracted from Ciao[2], a website specialized in user evaluations. We decided to use this site for its high-quality reviews, as users are paid to complete the tasks and because all of the texts are written in Spanish, the language of interest for our experiment.

Texts were annotated using Freeling[3] POS tagger (Padró et al. 2010). Then all of the units tagged as adjectives were extracted. We extracted the lemma of each adjective that appeared simultaneously in the three domains to compare their polarities. Finally, we obtained a list of 514 lemmas.

### 3.2 Polarity Annotation and Agreement

In order to determine the polarity of each lemma, five human annotators tagged each one. Annotation instructions were the following:

- -1 if the adjective is felt to be a negative feature, relative to the target domain.
- 0 if the adjective is felt to be irrelevant, not clearly related with a positive or negative feature to the target domain.
- 1 if the adjective is felt to be a positive feature, relative to the target domain.

All of the five human annotators were trained during two sessions to do the task. All of them had higher education level and were frequent users of review websites.

To measure the inter-annotator agreement we used the Kappa index (Siegel and Castellan, 1988). The values of the task were 0.59 for cars and mobile phones and 0.51 for films. These results showed that a domain such as films is more subjective (that is, annotators do not reach an agreement about the prior polarity of adjectives within this domain) than others, such as cars or mobile phones. These values gave us the first clue to think that some adjectives can have prior polarity variations not only across domains, but also within the same domain. Thus, we also used the standard deviation to measure this degree

---

[1] http://sentiwordnet.isti.cnr.it/
[2] http://www.ciao.es/
[3] http://nlp.lsi.upc.edu/freeling/

of variation within the same domain.

## 4. Data Analysis

The first aim was to calculate a value that could give us a measure of the polarity of each adjective. We calculated the arithmetic mean of the values assigned by the human annotators. If the human value was higher than 0, the element was considered to have a general tendency to be positive, while if this value was lower than 0, it was considered to have a general tendency to be negative. We also discovered some neutral adjectives which always (or in the majority of the cases) had a value equal to 0. Finally, the cases in which there were two -1, one 0 and two 1 were also considered as neutral elements. For instance, "adecuado" ("appropriate") had different values across the three domains but the arithmetical mean of all the tags is higher than 0, thus it tends to be positive. However, "aburrido" ("boring") has only negative values, thus it tends to be negative in any domain. Finally, "amarillo" ("yellow") was always annotated with neutral tags.

Some examples of the calculated arithmetic mean based on the annotation of the human annotators can be seen in Table 1.

| Adjective | Cars | Phones | Films |
|---|---|---|---|
| Adecuado | 0.5 | 0.5 | 0.5 |
| Aburrido | -1.0 | -0.75 | -1.0 |
| Amarillo | 0.0 | 0.0 | 0.0 |

Table 1: Arithmetic mean of "adecuado" ("appropriate"), "aburrido" ("boring") and "amarillo" ("yellow")

The principal aim of this assessment was to determine the general tendency of our adjectives to be in one or another side of the polarity scale and discover the elements that are neutral for some domains.

However, although the arithmetic mean showed us the general tendency of adjectives to be in a positive or negative group, we considered that it was not enough to uncover the real values of these elements. Therefore, we also calculated the standard deviation of all of the adjective tags over all domains and annotators. This value showed the dispersion or unification degree of opinions regarding the value of the arithmetic mean, besides the tendency of the adjective to be in a positive or a negative group.

With these results, we have considered as *domain independent* all of the units that were unanimously tagged as neutral (25.49%), negative (5.84%) or positive (1.36%), independently of domain and annotator.

We considered as *domain dependent* such units where tags change across domains.

We can see the results in Table 2.

For example, on the one hand, "decepcionante" ("disappointing) and "alucinante" ("amazing") are considered as negative and positive by all annotators, respectively. In other words, they are considered domain independent adjectives.

On the other hand, "pequeño" ("small") or "espacioso" ("spacious") are considered good features for some domains but bad for others, i.e. they are domain dependent. "Espacioso" ("spacious"), for example, is perceived as positive for cars but as an irrelevant feature for films and mobile phones and "pequeño" ("small") is positive for mobile phones, negative for cars and neutral for films.

| Type | Percentage |
|---|---|
| Domain independent | 32.68% |
| Domain dependent | 67.32% |

Table 2: Percentages of domain dependent and domain independent adjectives

Additionally, standard deviation from the arithmetic mean of all the annotations gave us the means to classify adjectives in three different categories: mixed, constant and highly subjective adjectives. We can see specific results in Table 3.

| Type | Percentage |
|---|---|
| Highly subjective | 22.18% |
| Mixed | 46.5% |
| Constant | 31.32% |

Table 3: Percentages of constant, mixed and highly subjective adjectives

*Highly subjective adjectives* have a high or very high standard deviation value in all the domains. This fact highlights the wide range of opinions about some adjectives from one annotator to another. Therefore, they are very subjective elements. There is no agreement about the annotation and they depend entirely on the subjectivity of each annotator. For example, "antiguo" ("old") is perceived as a good characteristic for films or cars by some annotator but as a bad characteristic by others. The specific annotation of "antiguo" ("old") can be seen in Table 4.

| Domain | Annotation |
|---|---|
| Cars | -1, -1, 1, -1, 1 |
| Phones | -1,-1,-1,0,0 |
| Films | 0,1,1,1,1 |

Table 4: Prior polarity annotations of "antiguo" ("old")

In the *mixed adjectives* group there are units with standard deviation values high or very high for a domain but with no deviation for other domains. These units are, obviously, domain dependent: in some cases, adjectives show a high subjective degree (that is, the annotators do not reach an agreement about its polarity) and in other cases the polarity of the adjective is clearly identified. An example is "agresivo" ("aggressive") in which does not exist agreement among the annotators in cars or films but has total agreement among the annotators for mobile phones (as neutral). The specific annotation of "aggressive" ("agresivo") can be seen in Table 5.

| Domain | Annotation |
|--------|------------|
| Cars | 0,1,0,1,-1 |
| Phones | 0,0,0,0,0 |
| Films | 0,0,0,0,-1 |

Table 5: Prior polarity annotations of "agresivo" ("aggressive")

The last group is made up of *constant polarity adjectives*, that is, adjectives whose standard deviation value is 0. In other words, annotators give the same tag to the adjective in all of the domains. In general, we could say that this group can be considered domain independent adjectives (except in the cases of "fiable" ("reliable") and "frágil" ("fragile"), which in the domains of cars and mobile phones are positive and negative, respectively while in films they are unanimously neutral). These adjectives (with these exceptions) could be automatically tagged with their semantic orientation because, their values do not change. Some examples are "bello" ("beautiful") or "pésimo" ("dreadful") that have the same tag among all of the annotators across all the domains. The specific annotation of "beautiful" ("bello") can be seen in Table 6.

| Domain | Annotation |
|--------|------------|
| Cars | 1,1,1,1,1 |
| Phones | 1,1,1,1,1 |
| Films | 1,1,1,1,1 |

Table 6: Prior polarity annotations of "beautiful" ("bello")

Some examples of standard deviation values can be seen in the Table 7.

| Adjective | Cars | Phones | Films | Type |
|-----------|------|--------|-------|------|
| Antiguo | 1.10 | 0.55 | 0.45 | Highly Subjective |
| Agresivo | 0.84 | 0.0 | 0.45 | Mixed |
| Bello | 0.0 | 0.0 | 0.0 | Constant |

Table 7: Standard deviation of "antiguo" ("old"), "agresivo" ("aggressive") and "bello" ("beautiful")

## 5. Conclusions

In this work we proved that there is variation in the annotation of the prior polarity of adjectives across domains. Our results indicate that a majority of adjectives are domain dependent elements and thus, we can not treat them as general units as it was done until now.

The advantage of domain dependent lexicons would be to refine textual analysis depending on the topic and to increase the precision of our sentiment analysis tools.

At present we are using all of the adjectives classified as domain independent elements as input of a bootstrapping methodology to automatically distinguish the semantic orientation of adjectives which prior polarity is unknown. Related to that, we are also working on the possibility of using these independent elements to increase the number of lexical entries that should be included in polarity lexicons, studying also interesting combinations like noun plus adjective.

Future work also includes the study of domain dependent elements with the intention of creating a classification of similar domains that can be studied together in order to develop polarity lexicons that could be shared by different domains.

## 6. Acknowledgements

This work was funded by the UPF-IULA PhD grant program.